\documentclass[Afour, sageh, times, doublespace]{sagej}
\usepackage{moreverb, url}
\usepackage[
    colorlinks,
    bookmarksopen,
    bookmarksnumbered,
    citecolor=red,
    urlcolor=red
]{hyperref}
\usepackage[acronym]{glossaries}
\usepackage{siunitx}
\usepackage{soul}
\usepackage{amsmath}
\usepackage{nicematrix}
\usepackage{dirtree}
\usepackage{orcidlink}
\usepackage{dblfloatfix}
\usepackage{float}
\usepackage{multirow}
\usepackage{caption}
\usepackage{subcaption}
\usepackage{courier}
\usepackage{comment}

\newcommand\BibTeX{{\rmfamily B\kern-.05em \textsc{i\kern-.025em b}\kern-.08em
T\kern-.1667em\lower.7ex\hbox{E}\kern-.125emX}}
\def\volumeyear{2025}

\newcommand{\etal}{\emph{et~al.}}
\newcommand{\secref}[1]{Section~\ref{#1}}
\newcommand{\figref}[1]{Figure~\ref{#1}}
\newcommand{\tabref}[1]{Table~\ref{#1}}

\begin{document}

\runninghead{Mansour \etal}
\title{eStonefish-Scenes: A Sim-to-Real Validated and Robot-Centric Event-based Optical Flow Dataset for Underwater Vehicles}

\author{Jad Mansour\affilnum{1}~\orcidlink{0009-0007-8559-6753}, Sebastian Realpe\affilnum{1}~\orcidlink{0000-0001-7823-5879}, Hayat Rajani\affilnum{1}~\orcidlink{0000-0002-2541-2787}, Michele Grimaldi\affilnum{2}~\orcidlink{0000-0001-8284-6998}, Rafael Garcia\affilnum{1}~\orcidlink{0000-0002-1681-6229}, and Nuno Gracias\affilnum{1}~\orcidlink{0000-0002-4675-9595}}

\affiliation{\affilnum{1}Computer Vision and Robotics Research Institute (ViCOROB), \\ Universitat de Girona (UdG), Spain \\ \affilnum{2}School of Engineering \& Physical Sciences, \\ Heriot-Watt University, Edinburgh, UK}

\corrauth{Jad Mansour, Edifici CIRS, Parc Científic i Tecnològic UdG, \\ Carrer Pic de Peguera 13, 17003 Girona, Spain}
\email{jad.mansour@udg.edu}

\newacronym{ebc}{EBC}{Event-based Camera}
\newacronym{uav}{UAV}{Unmanned Aerial Vehicle}
\newacronym{auv}{AUV}{Autonomous Underwater Vehicle}
\newacronym{rov}{ROV}{Remotely Operated Vehicle}
\newacronym{ann}{ANN}{Artificial Neural Network}
\newacronym{cnn}{CNN}{Convolutional Neural Network}
\newacronym{snn}{SNN}{Spiking Neural Network}
\newacronym{nn}{NN}{Neural Network}
\newacronym{if}{IF}{Integrate-and-Fire}
\newacronym{lif}{LIF}{Leaky Integrate-and-Fire}
\newacronym{aee}{AEE}{Average Endpoint Error}
\newacronym{aae}{AAE}{Average Angular Error}

\begin{abstract}
    Event-based cameras (EBCs) are poised to transform underwater robotics, yet the absence of labelled event-based datasets for underwater environments severely limits progress in tasks such as visual odometry and obstacle avoidance. Real-world event-based optical flow datasets are scarce, resource-intensive to collect, and lack diversity, while no prior benchmarks target underwater applications. To bridge this gap, we introduce eStonefish-Scenes, a synthetic event-based optical flow dataset generated using the Stonefish simulator, together with an open data generation pipeline for creating customizable underwater environments featuring realistic coral reefs and biologically inspired schools of fish with reactive navigation behaviours. We also present eWiz, a comprehensive library for event-based data processing, encompassing data loading, augmentation, visualization, encoding, training utilities, loss functions, and evaluation metrics.
    
    To validate sim-to-real transferability, we collected real-world data using a DAVIS346 hybrid event-and-frame camera mounted on a BlueROV2 in an indoor testing pool. Ground-truth optical flow was derived via homography-based frame-to-poster registration, and per-pixel uncertainty was estimated through Monte Carlo perturbation of keypoint correspondences. This uncertainty was incorporated into the evaluation metrics, enabling reliability-aware performance assessment. A ConvGRU-based optical flow network, trained exclusively on synthetic eStonefish-Scenes data, was evaluated on the real-world sequences without fine-tuning, achieving an uncertainty-weighted average endpoint error of 0.79 pixels. These results demonstrate that the proposed synthetic dataset effectively supports sim-to-real transfer for underwater event-based optical flow estimation, substantially reducing the need for costly real-world data collection.
\end{abstract}


\maketitle

\section{Introduction}
\label{sec:introduction}
    \Glspl*{ebc}, also known as dynamic vision sensors or neuromorphic cameras, are imaging sensors that operate differently from traditional frame-based cameras. Inspired by the motion perception of winged insects, these cameras respond asynchronously to brightness changes with microsecond temporal resolution \citep{camera1Lichtsteiner08, camera2Brandli14}. \Glspl*{ebc} offer significant advantages over traditional frame-based cameras: virtually no motion blur, high dynamic range, high temporal resolution, and low latency \citep{camera3Gallego20}. These unique properties make the \gls*{ebc} an ideal sensor to analyse dynamic scenes characterized by fast motion and rapid changes in lighting conditions.

    \Glspl*{ebc} and optical flow prediction are two complementary concepts that play an integral part in motion analysis in computer vision. Traditionally, optical flow prediction has relied on analysing a sequence of frames to infer motion patterns \citep{flowShah21}. However, \glspl*{ebc} present a new paradigm by providing an asynchronous and sparse representation of data in the form of a stream of events \citep{camera3Gallego20}. As such, many methodologies have been developed for event-based processing, aimed primarily at relatively simple scenes containing well-defined shapes, edges, and corners \citep{icpKueng16, probabilityZhu17, ekltGehrig20}. However, these methods have proven to be inefficient and computationally expensive in complex scenes \citep{trackingMessikommer23}.

    An alternative approach is the use of \glspl*{nn} for processing event-based data, which has demonstrated remarkable success in modelling and understanding the nature of biological vision systems. One significant drawback of \glspl*{ebc}, however, is their incompatibility with conventional \gls*{nn} models, such as \glspl*{cnn}. This requires complex data encoding schemes to convert the sparse asynchronous stream of events into a frame-based representation that can be handled by \glspl*{cnn}. This approach has been demonstrated by researchers \citep{evflownetZhu18, egoflownetZihao19, biflowWan22} who implemented U-Net-like architectures \citep{unetRonneberger15, unetWeng15} for event-based optical flow estimation. Compression of event streams into frame-based representations results in loss of information, thereby limiting the full potential of \glspl*{ebc}. As a result, there is a demand for newer paradigms specifically optimized for handling such data. An \gls*{snn} is one such brain-inspired computing model that relies on so-called spiking neurons to process information \citep{snnTavanaei19}. Spiking neurons are inherently designed to process temporal patterns, making them naturally compatible with the asynchronous stream of events. The interconnection of spiking neurons in an \gls*{snn} allows them to effectively extract spatial features and encode dynamic visual motion over time.

    Sufficiently diverse and well-labelled datasets to facilitate model training remain scarce. Generating real-world data is both expensive and time-consuming, often requiring specialized equipment to obtain corresponding ground truth data as an alternative to the tedious process of manual annotations by human experts. For instance, datasets such as MVSEC \citep{mvsecZhu18} and DSEC \citep{dsecGehrig21} utilize 3D LiDARs to create a 3D map of the environment. The reconstructed environment is then projected onto the \gls*{ebc} sensor to generate a dense optical flow for use as ground truth during training. However, the resulting optical flow is affected by clock synchronization issues and incomplete coverage, failing to fully represent the event sequences captured by the sensor. As such, these real-world datasets may not be ideal for model training or evaluation, which stresses the need for a synthetic alternative.

    Recently, the landscape of underwater event-based datasets has expanded with synthetic benchmarks such as UEOF \citep{ueofTruong26} and AquaticVision \citep{aquaticvisionPeng25}. UEOF provides a high-resolution benchmark derived from physically-based ray-traced RGBD sequences, utilizing a video-to-event (v2e) conversion pipeline to generate pseudo-events. However, this approach has notable limitations: it relies on post-processing pre-rendered video frames rather than natively integrating the event sensor within the simulator, and it requires manually crafted camera trajectories that are decoupled from any underlying robot dynamics. This decoupling prevents the dataset from capturing the authentic interplay between vehicle motion and sensor response characteristic of real underwater robotics scenarios. In contrast, our approach integrates the event camera directly into the Stonefish simulator, ensuring that the generated data captures the authentic coupling between the robot's hydrodynamics (e.g., heave, surge, sway) and the sensor's temporal response. Furthermore, unlike datasets that target high-resolution benchmarks, \emph{eStonefish-Scenes} deliberately retains the native resolution of the DAVIS346 sensor ($346 \times 260$ pixels). This design choice is critical for enabling the deployment of lightweight, resource-efficient neural networks capable of running on the power-constrained embedded hardware typical of AUVs. Finally, whereas existing benchmarks often lack a direct sim-to-real validation component, we validate our synthetic data against our own dedicated real-world dataset collected with a similar sensor configuration.
    
    In fact, due to the nature of \glspl*{ebc}, synthetic event-based data can translate well to real-world data. The features extracted from synthetic frame-based camera datasets often exhibit obvious differences from the features extracted from real-world data. These differences typically arise from distinct lighting conditions, camera noise, and environmental complexity in real-world settings, which are challenging to replicate accurately in synthetic data. Such variations can lead to shifts in pixel intensities, colour distribution and texture details, resulting in significant differences between the features extracted from synthetic and real-world images. In contrast, \glspl*{ebc} only trigger events when detecting brightness changes, producing an event representation that only has two values: a positive or a negative polarity. As such, unlike frame-based images that rely on a static grid of pixels with values typically ranging between 0 and 255, the output of \glspl*{ebc} is simpler, allowing for greater tolerance to domain differences when transitioning from simulation.
    
    To the best of our knowledge, no event-based optical flow prediction datasets currently exist for underwater environments. Capturing such data in real-world conditions presents significant challenges due to several factors, such as poor lighting conditions, water turbidity, and light scattering. These factors adversely affect the performance of conventional camera systems typically used for capturing optical flow data. To overcome these limitations, we take a synthetic approach, enabling precise control over environmental conditions and ensuring reliable ground truth. A synthetic event-based dataset for underwater optical flow prediction opens new avenues for research on \glspl*{ebc} in these conditions. Light absorption and scattering, especially at greater depths, reduce illumination and cause colour distortion. Unlike conventional cameras, \glspl*{ebc} respond to changes rather than intensities of ambient light, allowing them to function effectively in low-light scenarios. Moreover, \glspl*{ebc} are resilient to motion blur and noise caused by scattering effects, making them ideal for capturing fast-moving objects in dynamic underwater environments.
    
    To address this need, we present \emph{eStonefish-Scenes}, a synthetic event-based optical flow dataset tailored specifically for underwater environments, together with an easy-to-use pipeline for creating custom datasets. \emph{eStonefish-Scenes} is based on the Stonefish simulator \citep{stonefishCielak19, stonefishGrimaldi25}, with a \gls*{rov} capturing event-based, grayscale, and optical flow data. To enhance the realism of the simulations, we introduce \emph{Stonefish-SceneGen}, a scene generator capable of randomly populating the environment with corals, creating rich and vibrant lifelike reefs. To animate these scenes, we also present \emph{Stonefish-Boids}, a package that simulates the behaviour of schools of fish. Together, these components enable the creation of realistic underwater scenarios. The dataset is hosted on Zenodo (\url{https://doi.org/10.5281/zenodo.15130452}) and the data generation pipeline is publicly available at: \url{https://github.com/CIRS-Girona/stonefish-scenegen}.

    We also introduce \emph{eWiz}, a library comprising of all the tools necessary for processing event-based data, ranging from data loading and manipulation, to augmentation and visualization, to encoding and training data generation, together with various loss functions and performance metrics. The library is openly available at the following link: \url{https://github.com/CIRS-Girona/ewiz}.

    To evaluate the sim-to-real transfer performance of the synthetic dataset, a real-world experimental study was conducted. A DAVIS346 event-based camera was rigidly mounted on a BlueROV2 platform and deployed in the indoor testing pool at the Centre for Research in Underwater Robotics and Vision (CIRS) in Girona. To enable controlled and repeatable optical flow estimation, a large, high-resolution textured poster was placed on the pool floor, providing a planar reference surface with rich visual features. Ground-truth optical flow was computed via frame-to-poster feature matching using a homography-based registration approach \citep{homoHartley04}. To quantify the reliability of the resulting flow, per-pixel uncertainty was estimated via Monte Carlo perturbations of keypoint correspondences, yielding covariance-based uncertainty maps aligned with the optical flow field. This uncertainty information was subsequently incorporated into the evaluation metrics, enabling confidence-weighted performance assessment rather than relying on fixed outlier thresholds commonly used in optical flow benchmarks. The real-world validation dataset for \emph{eStonefish-Scenes} can be found on Zenodo (\url{https://doi.org/10.5281/zenodo.18473013}).

    Finally, sim-to-real generalisation was assessed by implementing a ConvGRU-based event-based optical flow network inspired by \citep{firenetHagenaars21}. Training was performed exclusively on the synthetic \emph{eStonefish-Scenes} dataset and evaluation on the real-world sequences. The network was trained using self-supervised photometric and smoothness losses defined over static scenes \citep{evflownetZhu18}, and its performance on real data was measured using uncertainty-aware error metrics. This evaluation demonstrates the effectiveness of the proposed synthetic dataset for learning transferable motion representations and highlights its suitability for underwater event-based vision tasks.

    The main contributions of the work are:
    \begin{itemize}
        \item The creation of \emph{eStonefish-Scenes}, an event-based optical flow dataset specifically tailored for underwater environments, along with an open data generation pipeline that simplifies the process of extending the dataset. To the best of our knowledge, no prior event-based datasets exist for underwater applications.
        \item The development of \emph{Stonefish-SceneGen}, a scene generator for Stonefish to streamline creation of diverse, coral-rich underwater environments.
        \item The development of \emph{Stonefish-Boids}, a package for simulating schools of fish in Stonefish, enabling the generation of dynamic underwater scenes.
        \item The development of \emph{eWiz}, a comprehensive framework for event-based data processing and manipulation.
        \item A real-world underwater data acquisition approach using an event-based camera mounted on an ROV, along with a homography-based method for ground-truth optical flow computation and per-pixel uncertainty estimation.
        \item An uncertainty-aware evaluation framework for event-based optical flow, incorporating covariance-based confidence measures into standard performance metrics.
        \item A sim-to-real validation study demonstrating the effectiveness of the proposed dataset by training a ConvGRU-based event-based optical flow network exclusively on synthetic data and evaluating it on real underwater sequences.
    \end{itemize}

    The remainder of this paper is organized as follows. \secref{sec:related_work} reviews publicly available event-based datasets for optical flow prediction. \secref{sec:simulation_setup} describes the simulation setup and sensor configurations used to generate \emph{eStonefish-Scenes}, while \secref{sec:data_generation} details the data generation and collection pipeline. \secref{sec:data_processing} presents the \emph{eWiz} library for event-based processing, visualization, encoding, and training utilities. \secref{sec:dataset_structure} describes the dataset structure and storage format. \secref{sec:ground_truth_acquisition} introduces the real-world validation dataset acquisition methodology and the homography-based ground-truth optical flow computation with uncertainty estimation. \secref{sec:sim-to-real_evaluation} evaluates sim-to-real transfer by training an event-based optical flow network on synthetic data and testing on real sequences. Finally, \secref{sec:conclusion} concludes the paper and outlines future work.

\section{Related Work}
\label{sec:related_work}
    In the literature, only a limited number of real-world event-based datasets have been employed for developing and evaluating event-based vision algorithms and \glspl*{nn} targeting optical flow prediction. Among these, the widely used MVSEC dataset \citep{mvsecZhu18} stands out, offering two distinct scenarios: a stereo pair of \glspl*{ebc} mounted on a \gls*{uav} in an indoor environment, and the same setup mounted on a car in an outdoor setting. The ground truth optical flow for indoor scenes is derived using a motion capture system, while for outdoor scenes, it is obtained through a combination of GPS and scene reconstruction techniques. Building upon MVSEC, the DSEC dataset, introduced by \citet{dsecGehrig21}, presents significant improvements, including higher camera resolution and larger displacement fields. While these datasets have been instrumental in advancing event-based vision research, they are not without limitations. The generation of real-world data is both resource-intensive and time-consuming, often requiring specialized equipment to capture accurate ground-truth information. The MVSEC dataset exhibits notable limitations in the accuracy of its ground-truth optical flow. The ground truth was generated using a Velodyne LiDAR, which relies on a 3D reconstruction of the environment to estimate optical flow. The dataset suffers from additional challenges due to the mechanical setup. Specifically, the vibrations of the car during forward motion induce a wobbling effect in the event-based sensor. This wobbling motion leads to the sensor capturing significant vertical motion, whereas the ground truth optical flow predominantly reflects forward motion. This mismatch between the motion perceived by the event-based sensor and the ground-truth optical flow can introduce discrepancies when training and evaluating models, especially for applications that rely on precise motion estimation.

    These challenges contribute to the scarcity of real-world event-based datasets, prompting researchers to increasingly turn to synthetically generated datasets, produced by advanced simulators. ESIM \citep{esimRebecq18}, for example, being one of the most accurate in the literature, makes use of an adaptive rendering scheme that only samples frames when necessary. Such a rendering scheme adapts the sampling rate to the predicted dynamics of the visual signal, allowing for a faithful simulation of events when fast motion occurs. ESIM has been used to generate several synthetic event-based optical flow datasets, including MultiFlow \citep{multiflowGehrig24} and HREM \citep{hremLuo24}. MultiFlow contains multiple moving objects undergoing continuous similarity transformations, along with dense ground-truth optical flow. HREM, on the other hand, comprises 100 virtual scenes designed to replicate real-world indoor and outdoor environments. BlinkSim, proposed by \citet{blinksimLi23}, is another \gls{ebc} simulator that includes a powerful rendering engine capable of generating thousands of scenes with different objects and motion patterns at high frequency. This simulator gave rise to the BlinkFlow dataset, which improves the generalization performance of state-of-the-art methods by 40\% on average. Another synthetic dataset is MDR created by \citet{mdrLuo23}. It makes use of Blender \citep{blenderBlender23} to create indoor and outdoor 3D scenes and includes diverse camera motions and high-frame-rate videos.

    One of the key reasons behind generating synthetic datasets is the ability to diversify the created environments and easily extend applications related to \glspl*{ebc}. Although synthetic optical flow datasets for event-based vision are already available, most are not specifically designed to reflect the motion dynamics or characteristics of any particular vehicle type. MDR, MultiFlow, and BlinkFlow are limited to predefined camera trajectories, which restricts their ability to accurately replicate such dynamic motion patterns. Furthermore, these simulators lack sophisticated light behaviour modelling, which is essential for creating realistic and complex environments. Phenomena such as light absorption, scattering, and refracted sunlight are key characteristics of underwater environments that these simulators fail to account for, thus limiting the fidelity of synthetic datasets for underwater applications. On the other hand, ESIM makes use of an OpenGL rendering engine \citep{openglOpengl14}, which requires manual development of every rendering pipeline and shader from scratch, making it even more complex for underwater environments. Similarly, the use of Blender \citep{blenderBlender23} for MDR requires some knowledge in 3D scene creation to appropriately simulate realistic conditions, especially for underwater environments. All these factors make the creation of synthetic event-based datasets a challenging endeavour, particularly for underwater environments.
    
    Currently, research on underwater event-based applications is highly limited, especially due to the lack of publicly available datasets captured with \glspl*{ebc} in underwater environments. To address this gap, we created a diverse synthetic dataset tailored for underwater applications, along with a modular data generation pipeline to facilitate the creation of custom datasets. The dataset incorporates both static and dynamic elements, enabling the trained algorithms to learn robust feature representations by capturing background stability and motion-driven variations. This also enhances generalization to real-world scenarios, where stationary structures and moving entities, such as marine life, coexist.

    Finally, although a few open-source tools are available for specific event-based data processing tasks, such as the work by \cite{mcShiba2022}, no comprehensive library currently integrates these capabilities into a unified framework. To address this gap and streamline future development, we developed \emph{eWiz}, a comprehensive library for event-based data manipulation and processing.

\section{Simulation Setup}
\label{sec:simulation_setup}
    This section describes the simulation environment, sensor and vehicle configurations, and the data collection approach for generating the \emph{eStonefish-Scenes} dataset. The dataset consists of event streams, grayscale images, and the corresponding ground-truth optical flow. Additionally, it utilizes an optimized format provided within the \emph{eWiz} framework to facilitate efficient data storage, access, and processing. Further details on this optimized format are provided in \secref{sec:data_processing}.

    The dataset was generated using the Stonefish simulator \citep{stonefishCielak19, stonefishGrimaldi25} with a custom-designed environment featuring a texture-rich seabed. To add visual complexity, the seabed was populated with abundant corals and marine flora featuring pronounced edges and contrasts. The corals were divided into 3 clusters distributed across the seabed, with each cluster containing 20 corals of different sizes, ranging from small branching corals to large dome corals. The scene was randomly generated using a scene generator to diversify coral poses and sizes. The dataset features two environments:
    \begin{itemize}
        \item \textbf{Rocky:} Sequences characterized by a barren rocky seabed with no flora in the scene.
        \item \textbf{Reef:} Sequences characterized by the same rocky seabed texture, but populated with flora to simulate the look and feel of a colourful coral reef.
    \end{itemize}
    
\begin{figure}[!h]
        \centering
            \includegraphics[width=0.4\textwidth]{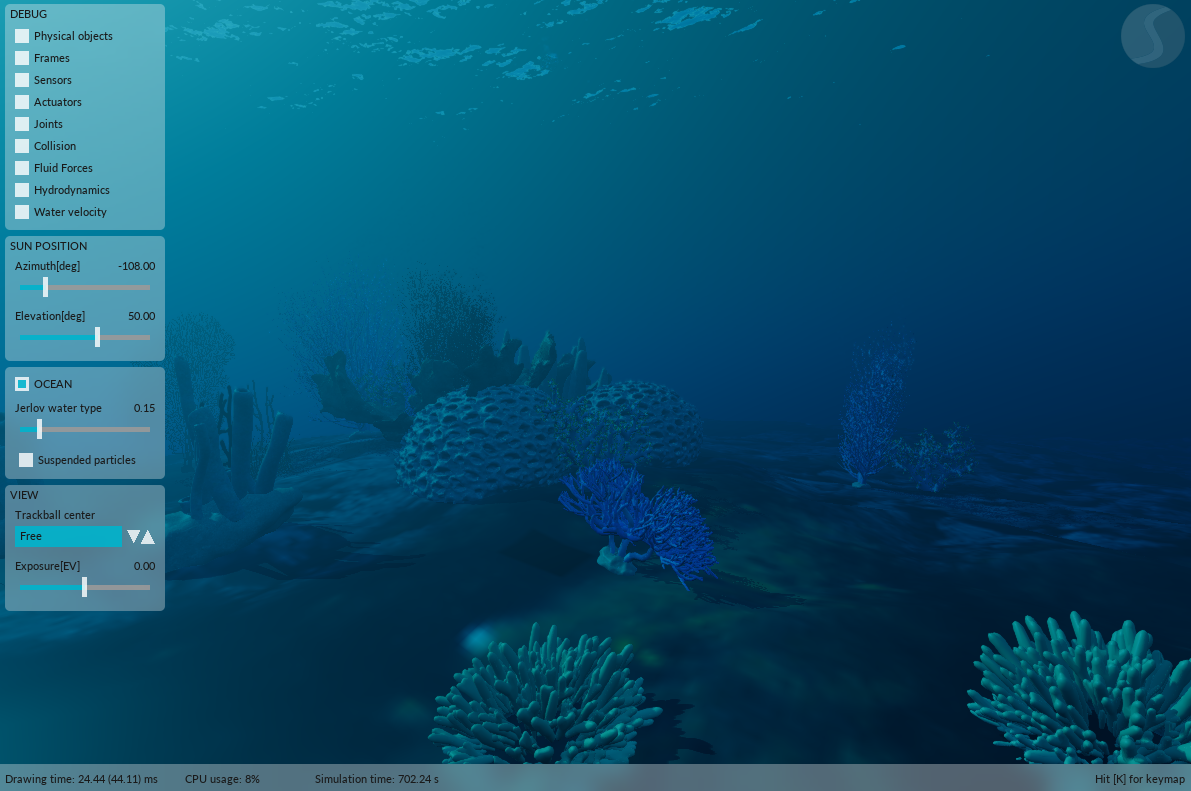}
            \caption{Generated environment: the environment includes a seabed with feature-rich corals.}
        \label{fig:coral_environment}
    \end{figure}

    The dataset comprises both static and dynamic scenes. Static scenes are simple environments with no dynamic objects, such as schools of fish. Only stationary corals appear in these scenes, ensuring consistency with event-based optical flow loss functions commonly utilized in the literature \citep{mcStoffregen19}. Dynamic scenes, on the other hand, incorporate moving elements in the form of schools of fish. Several fish models were used across the simulation, each exhibiting slightly modified Boid behaviours \citep{boidsReynolds87} to mimic natural schooling behaviour of fish in nature. This diversity in scene complexity, which creates complex visual patterns, provides a comprehensive framework for both training and evaluating optical flow algorithms across various environments. \figref{fig:coral_environment} depicts an example of the generated scene.

\begin{figure}[!h]
        \centering
            \includegraphics[width=0.4\textwidth]{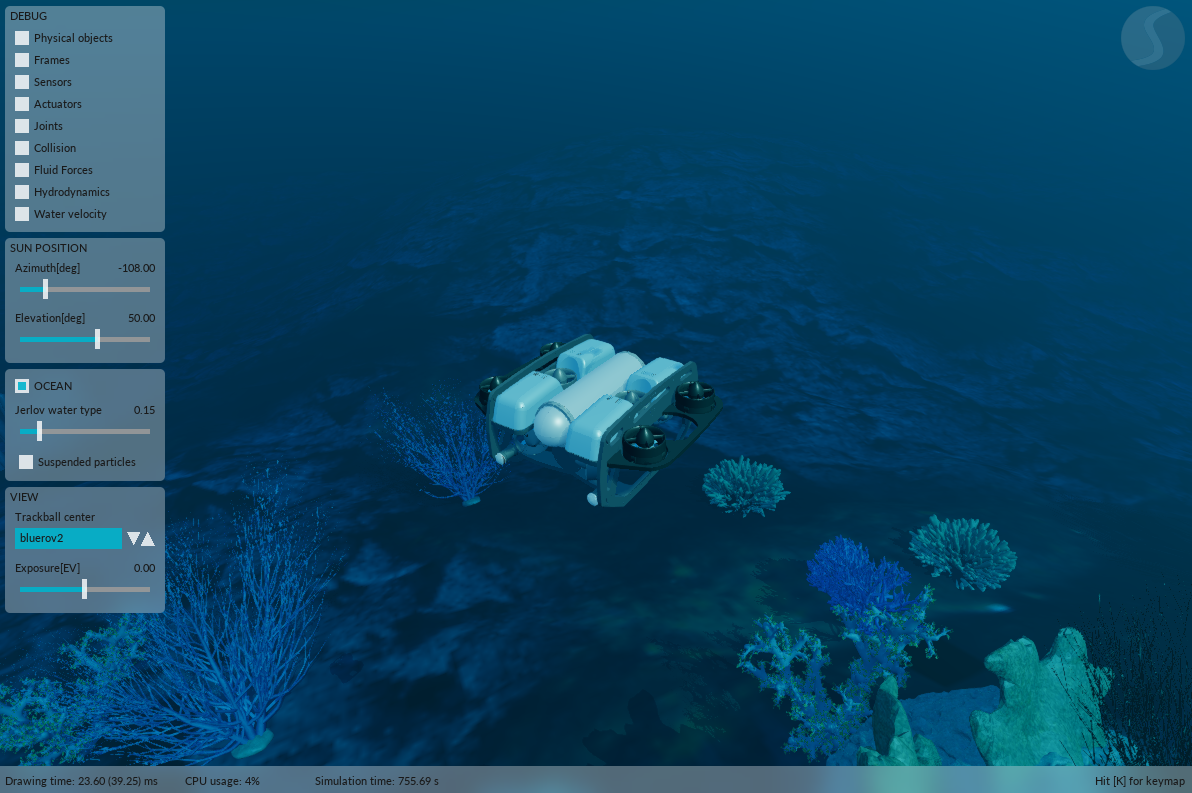}
            \caption{Simulated BlueROV2 navigating an underwater environment.}
        \label{fig:bluerov2_robot}
    \end{figure}

    For data collection, a joystick-controlled \emph{BlueROV2} \citep{bluerov2} model was utilized, as shown in \figref{fig:bluerov2_robot}. The \gls*{rov} was equipped with an RGB camera, an event-based camera, and an optical flow sensor, all of which are mounted in two configurations. Unlike the RGB and EBC cameras, that are modelled after real sensors in Stonefish, the optical flow sensor is a pseudo-sensor that uses the information from the 3D environment to create a perfect, noise-free representation of the optical flow data. To ensure diverse and robust data, the scene was recorded from multiple viewpoints, mimicking the placement of a camera on a real \gls*{rov}. Specifically, the sensors were positioned in the following configurations:
    \begin{itemize}
        \item \textbf{Forward-looking:} In this setup, the depth of the \gls*{rov} was adjusted dynamically while moving.
        \item \textbf{Down-looking:} In this setup, the \gls*{rov} maintained a constant depth while moving, focusing on the seabed.
    \end{itemize}
    
    The forward-looking configuration is particularly advantageous for tasks such as scene inspection and obstacle avoidance, as it provides a clear view of the environment ahead. In contrast, the down-looking configuration is optimized for computing the robot odometry, leveraging the \gls*{ebc} feed to analyse the textured seabed. These complementary configurations enable the \gls*{rov} to address a diverse range of underwater exploration and navigation challenges.
    
    All cameras mounted on the \gls*{rov} have a resolution of $346\times260$ pixels and a field-of-view of \SI{70}{\degree}. The RGB, EBC, and optical flow sensors share the same intrinsic and extrinsic camera parameters, and can be considered as a single camera that produces perfectly aligned colour, event, and optical flow data. The \gls*{ebc} was configured with positive and negative thresholds of 0.28, and a refractory period of \SI{200}{\nano\second}. The simulation clock was set to \SI{300}{\hertz}. An NVIDIA RTX 4090 GPU was used for the simulation and data collection. This configuration capped the grayscale and optical flow sensors at \SI{20}{\hertz}, while the event-based camera compared the frame brightness at 17 FPS to maintain real-time simulation. Nonetheless, events are triggered based on the defined contrast threshold and refractory period to ensure that the temporal resolution of events adapts dynamically to scene changes. When the intensity difference at a pixel exceeds an integer multiple of the contrast threshold, a corresponding number of events are generated with evenly distributed timestamps between the 2 frames where that intensity difference was computed. This preserves the temporal granularity of the events, as it produces multiple events during that interval.
    
\begin{figure*}[!h]
        \centering
            \includegraphics[width=1\textwidth]{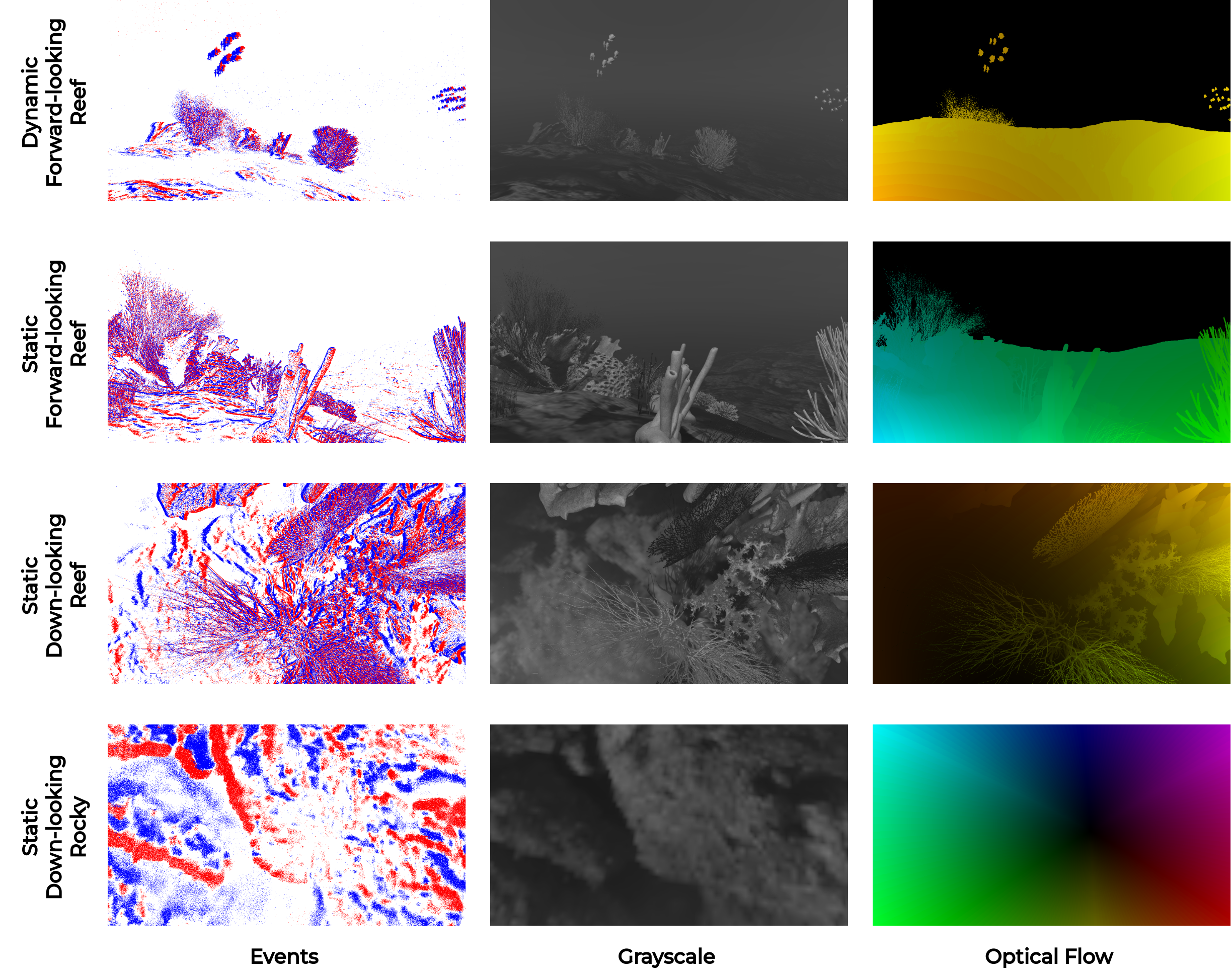}
            \caption{Visualisations of event-based, grayscale, and optical flow images for different generated sequences. Event-based images are accumulated over \SI{100}{\milli\second} intervals.}
        \label{fig:dataset_table}
    \end{figure*}

    Each collected sequence is \SI{45}{\second} long, during which the \gls*{rov} moves and rotates in all degrees of freedom (forward, backward, up, down, and sideways) to capture diverse flow intensities and spatial distributions. These motions are performed at varying speeds up to a maximum velocity of \SI{0.85}{\meter/\second}. A total of \textbf{10} sequences were collected. All sequences, along with their corresponding setups, are summarized in \tabref{table:sequences_list}. A sample visualization of the dataset is shown in \figref{fig:dataset_table}.
    \begin{table*}[!h]
        \centering
        \begin{tabular}{||c|c|c|c|c||}
            \hline
            Sequence Number & Scene Type & Altitude & Environment & Camera Position \\
            \hline\hline
            1 & Static & Constant & Reef & Down-looking \\
            \hline
            2 & Static & Constant & Rocky & Down-looking \\
            \hline
            3 & Static & Varying & Reef & Down-looking \\
            \hline
            4 & Static & Varying & Reef & Front-looking \\
            \hline
            5 & Static & Varying & Rocky & Down-looking \\
            \hline
            6 & Static & Varying & Rocky & Front-looking \\
            \hline
            7 & Dynamic & Varying & Reef & Down-looking \\
            \hline
            8 & Dynamic & Varying & Reef & Front-looking \\
            \hline
            9 & Dynamic & Varying & Rocky & Down-looking \\
            \hline
            10 & Dynamic & Varying & Rocky & Front-looking \\
            \hline
        \end{tabular}
        \caption{Summary of the generated sequences and their corresponding configurations in the \emph{eStonefish-Scenes} dataset.}
        \label{table:sequences_list}
    \end{table*}

    These sequences can be utilized for training neural networks by dividing them into separate event volumes. Each event volume contains a stream of events, along with all grayscale images acquired within the corresponding timestamps. One such example is extracting events that occur between every pair of consecutive grayscale images. The resulting volumes include a grayscale frame at time $t$, another frame at time $t + 1$, and all events occurring in between. Based on the 20 FPS refresh rate of the grayscale sensor, we can extract approximately 800 training samples per sequence, providing around 6,400 samples for all sequences combined.

\section{Data Generation Pipeline}
\label{sec:data_generation}
    The dataset can be easily extended using a modular data generation pipeline, which comprises three separate packages, each responsible for different aspects of data generation. First, we present the \emph{Stonefish-SceneGen} package. This package is used to generate coral-rich and vibrant scenes by randomly populating the seabed with corals of different types and sizes. It generates a Stonefish-compatible scene file, which can then be used to launch the main pipeline, \emph{eStonefish-Scenes}. The pipeline is a ROS package with a launch file for Stonefish. The launch file starts the simulator with the generated environment, spawning the \gls*{rov} and displaying live feeds from all its sensors. The pipeline internally uses the \emph{Stonefish-Boids} package to randomly spawn schools of fish in the scene. The user can then move the \gls*{rov} around the scene while recording a rosbag, which can subsequently be converted to the \emph{eWiz} format using the \emph{ConvertStonefish} module of \emph{eWiz}.

    \subsection{Scene Generation}
        One challenge with the Stonefish simulator is the time-consuming process of manually creating rich and dynamic underwater scenes. Each coral or 3D object must be defined in a scene file, specifying its pose in the world frame, size, and physical attributes. While this approach is feasible for a few objects, it becomes impractical for environments featuring numerous objects. To address this limitation, we introduce \emph{Stonefish-SceneGen}, an underwater scene generator specifically designed for the Stonefish simulator. This tool streamlines the creation of complex underwater environments, significantly reducing the effort required for scene generation. It consists of a Python script that takes as input a Stonefish scene file, a seafloor surface, and the desired 3D models to spawn. The user specifies the number of coral clusters, along with the number of corals spawned in each cluster. Each coral instance is placed by sampling a random vertex of the seafloor surface. The surface normal at that vertex is then used to determine translation and rotation transformations. To ensure reproducibility, the user can generate the exact same scene by fixing a random seed before running the script. Finally, a scene file containing the path, texture, size, and pose of all spawned 3D models is generated. This scene file can then be used with \emph{eStonefish-Scenes} to launch the simulation. The scene generator is made openly available at the following link: \url{https://github.com/CIRS-Girona/stonefish-scenegen}.

    \subsection{Simulating Schools of Fish}
        To enhance the visual complexity of the underwater scenes with dynamic elements, we incorporated an adapted implementation of the Boids algorithm \citep{boidsReynolds87} for simulating the natural grouping behaviour observed in schools of fish. The algorithm is based on four fundamental rules that govern the behaviour of each fish within the school, collectively influencing group behaviour as a whole.
        \begin{itemize}
            \item \textbf{Alignment:} Each fish adjusts its velocity to match the average velocity vector of nearby fish within a defined perception radius.
            \item \textbf{Cohesion:} Each fish steers toward the average position of nearby fish within its perception radius.
            \item \textbf{Separation:} Each fish receives a repulsion force if it gets too close to another fish in the school, maintaining a minimum distance between individuals.
            \item \textbf{Leader Attraction:} One fish is designated as the leader of the school and follows a predefined path through the environment, while the other fish experience an attractive force toward the leader.
        \end{itemize}

        At each simulation step, the total force acting on each fish is calculated as a weighted sum of the contributions from each rule, which is then used to derive the change in position. This force-based approach can be extended beyond the four fundamental behavioural rules, allowing additional forces to dynamically influence movement in response to the surroundings. For instance, external forces such as repulsion from obstacles help maintain realistic movement patterns by preventing collisions with elements like corals, miscellaneous structures, the water surface and the \gls*{rov}.

\begin{figure}[h]
            \centering
            \begin{subfigure}[b]{0.4\textwidth}
                \centering
                \includegraphics[trim={0 0 0 0}, clip, width=\linewidth]{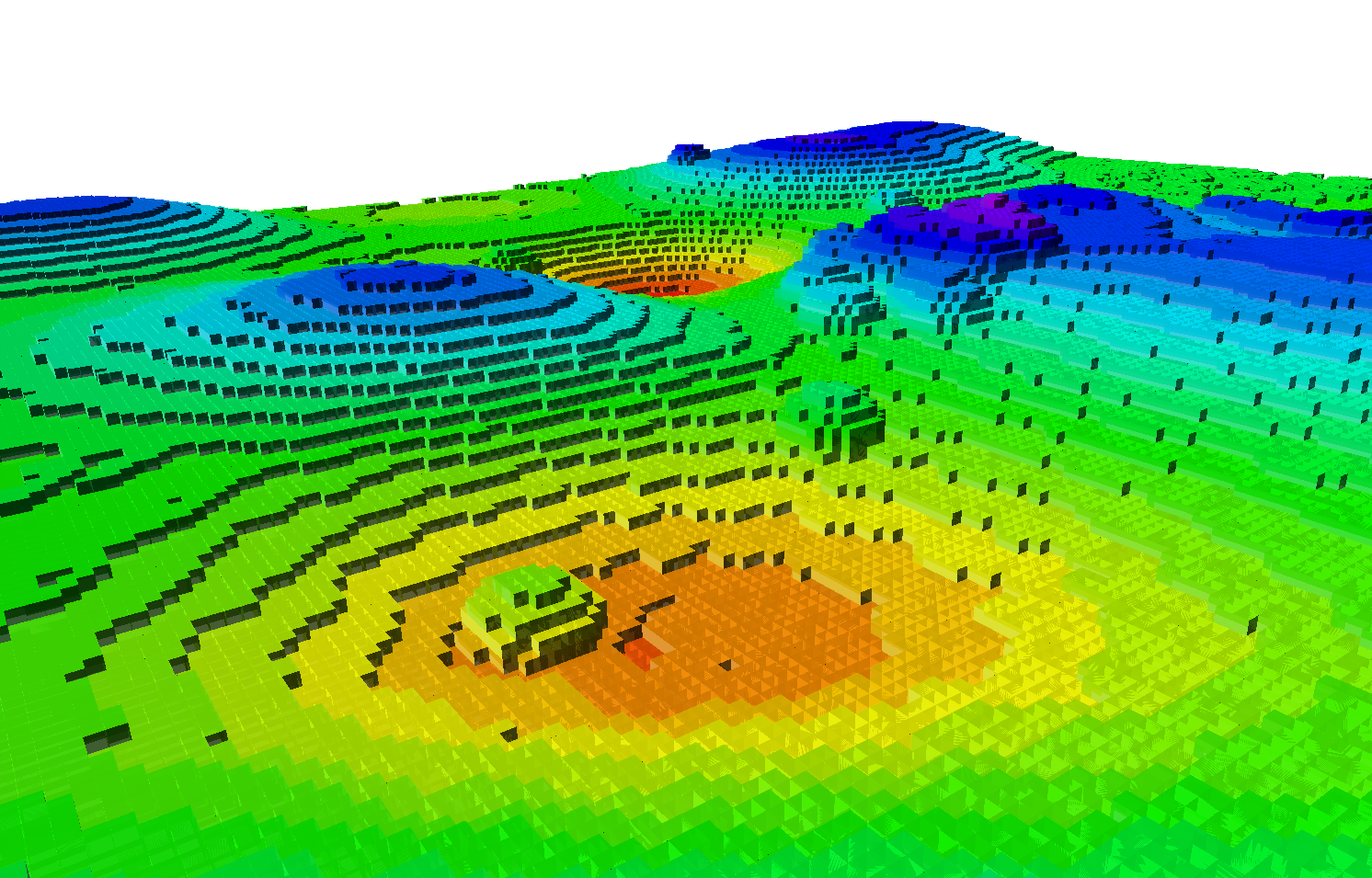}
                \captionsetup{justification=centering}
                \caption{Octree Scenario}
            \end{subfigure}
            \begin{subfigure}[b]{0.4\textwidth}
                \centering
                \includegraphics[trim={200 150 200 100}, clip, width=\linewidth]{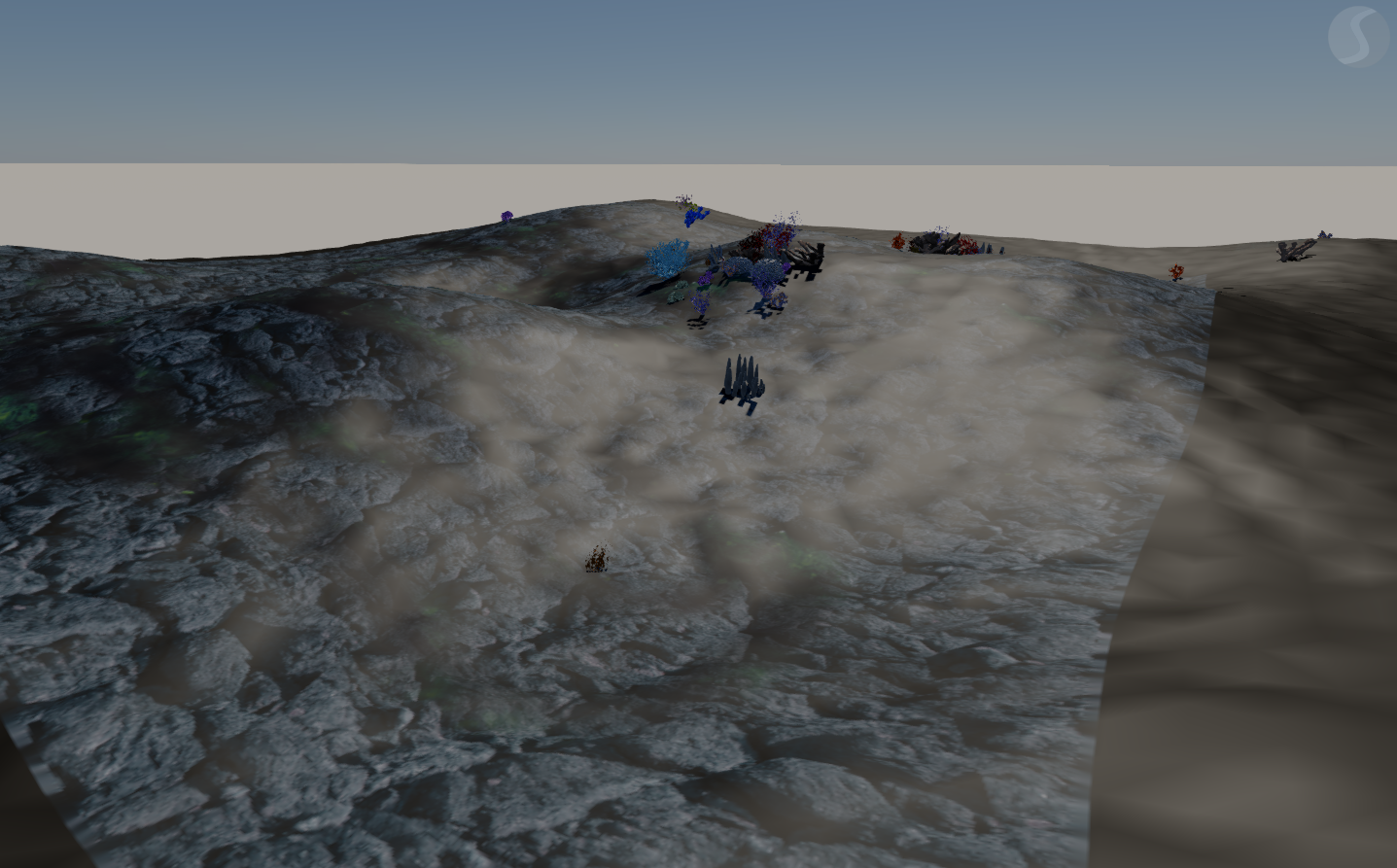}
                \captionsetup{justification=centering}
                \caption{Seabed and Corals Scenario}
            \end{subfigure}
            \caption{Discretised underwater environment represented as an Octree structure.}
            \label{fig:octomap}
        \end{figure}

        To efficiently handle such environmental interactions with static objects, we use OctoMap \citep{octomapHornung13} for discretizing the simulation space, as illustrated in Figure \ref{fig:octomap}. This enables rapid detection of potential collision zones and the assignment of repulsive force vectors to different locations. The flexibility of the library allows for the creation of a custom Octree in which each node stores a repulsive force vector normal to the object surface. When a fish enters an obstacle voxel, a repulsive force is applied, ensuring smooth and natural avoidance behaviour. For dynamic objects such as the \gls*{rov}, repulsive forces are applied based on the distance between each fish and a defined repulsion radius around the object, allowing for real-time interaction.

        Additionally, each school of fish follows a randomly generated path created using the Rapidly-exploring Random Trees (RRT) algorithm \citep{rrtSteven98}, implemented through the Open Motion Planning Library (OMPL) \citep{omplSucan12}. The designated leader follows this path, while the rest of the group moves accordingly based on the leader attraction rule. Each time a new simulation is initiated, both the path and the spawning positions of the fish are randomly generated. However, for reproducibility, users can set a fixed seed to ensure the path and spawning positions remain consistent across runs.

        Users can customize the type of fish schools and adjust various behavioural parameters through a YAML configuration file. These include alignment, cohesion, and separation weights, as well as spawning location, model scale, and school size. This functionality is provided as an open-source ROS package, \emph{Stonefish-Boids}, which builds on the base Stonefish API. The package is available at: \url{https://github.com/CIRS-Girona/stonefish_boids}.

    \subsection{Data Collection}
        To record data from the generated scenes, we provide a ROS-based utility, \emph{eStonefish-Scenes}, which uses the scene file generated by the scene generator and internally calls \emph{Stonefish-Boids} to simulate schools of fish. It relies on the Stonefish-ROS bridge to facilitate communication with the simulator. The package also includes meshes and textures for the BlueROV2 and a rocky seabed. A predefined scenario combines these assets into a single environment, placing the \gls*{rov} and the seabed beneath. A Python script is provided to control the robot, utilizing its allocation matrix to enable joystick-based operation. Additionally, an events rendering script is included to visualize the resulting event image in RViz \citep{kamRviz15} as the simulation progresses. Users can customize the sensors mounted on the \gls*{rov} by modifying the corresponding scene file. Customization options include, but are not limited to, adjusting resolution, field of view, event-based sensor thresholds, as well as adding noise. The \emph{record.bash} script allows users to record rosbags from the simulation, which can subsequently be converted to the \emph{eWiz} format using the \emph{ConvertStonefish} module of \emph{eWiz}. This data collection utility is available at the following link: \url{https://github.com/CIRS-Girona/estonefish-scenes}.

\section{Data Processing}
\label{sec:data_processing}
    The \emph{eStonefish-Scenes} dataset was generated with \gls*{nn} training and inference in mind. To streamline this process, we created \emph{eWiz}, a Python-based library that can be installed with PyPi \citep{pipyInstaller}, serving as an all-inclusive framework for event-based data manipulation, processing, visualization, and evaluation. \emph{eWiz} is designed for seamless integration with libraries such as PyTorch \citep{pytorchLibrary} and Tonic \citep{tonicLibrary}, making it suitable for machine learning-oriented pipelines. Additionally, it offers compatibility with existing datasets and supports both event-based and frame-based data. The following subsections present an overview of the most relevant functionalities of \emph{eWiz}. The library is openly available at the following link: \url{https://github.com/CIRS-Girona/ewiz}.
    \begin{figure}[!h]
        \includegraphics[width=0.45\textwidth]{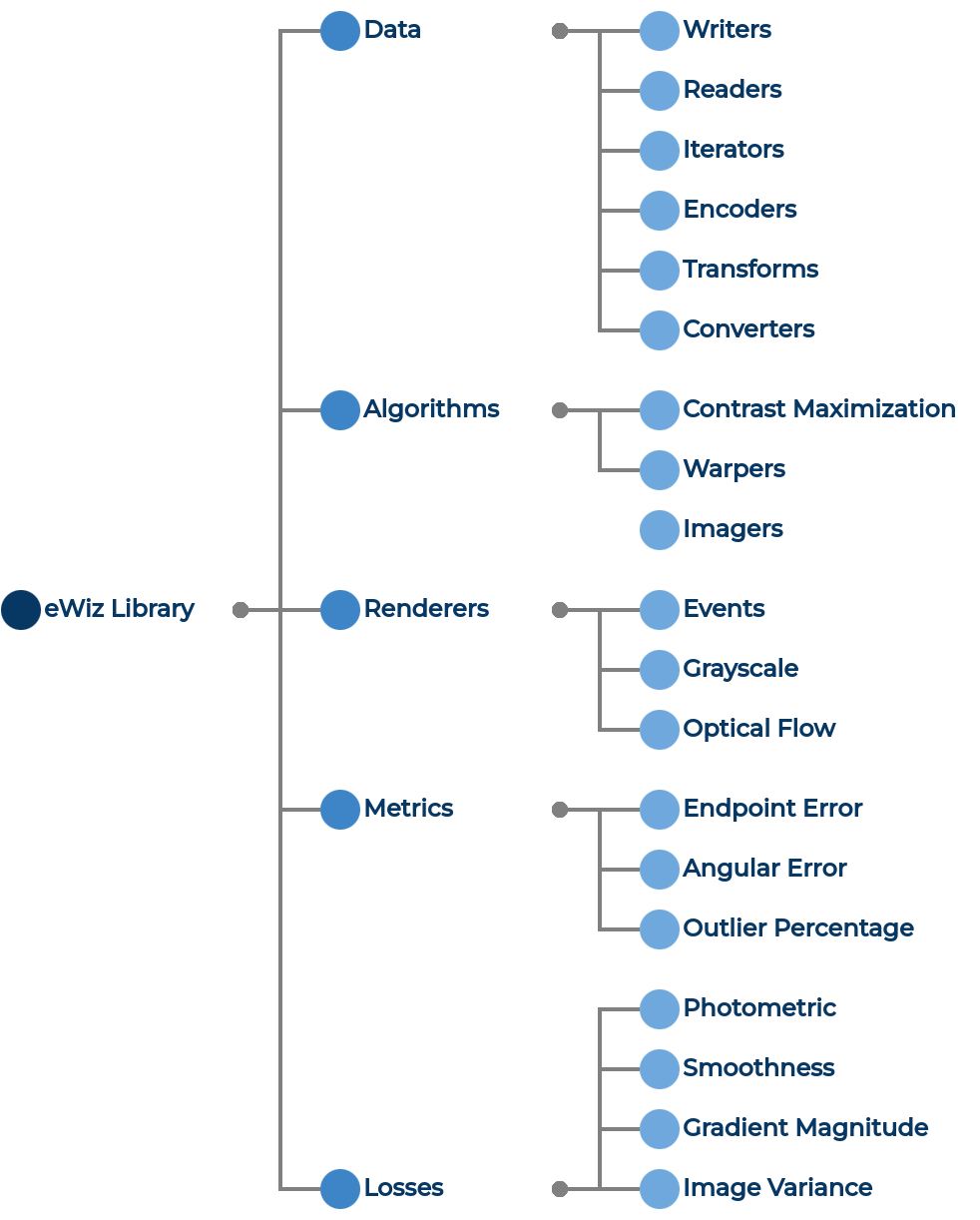}
        \caption{Overview of currently implemented modules in the \emph{eWiz} library.}
        \label{fig:ewiz_modules}
    \end{figure}

    \subsection{Data Access and Storage}
        Due to the high temporal resolution and data throughput of \glspl*{ebc}, event-based data requires substantial storage space. Additionally, slicing data between any two timestamps requires time-consuming search algorithms due to the large volume of events in the sequence. Libraries like NumPy \citep{numpyNumpy} also face high memory demands, as they typically require loading entire sequences into memory before slicing or processing. The \emph{eWiz} format addresses these challenges with an optimized approach to data management and manipulation. First, we use Blosc \citep{bloscCompression}, a high-performance compressor optimized for binary data and effective for compressing numerical arrays. To further reduce data size, the possible values for every event stream was constrained. For instance, pixel coordinates are stored as 16-bit unsigned integers, while polarities are stored as Boolean values. Next, we use the \emph{h5py} library \citep{h5pyLibrary} to read and write data as HDF5 files \citep{hdf5Format}, which support large and complex data structures, and allow slicing without the need to completely load the data into memory. Finally, to reduce loading times, the mappings between timestamps and data indices were pre-computed during the data writing process. This includes binary search to map timestamps (in $ms$) to event, grayscale, and flow indices, as well as direct mappings of event indices to their grayscale and flow counterparts. Additionally, optical flow data is automatically synchronized between specified timestamps by interpolating and accumulating flow displacements as needed.

        \emph{eWiz} provides easy-to-use readers and writers to facilitate this process. The data reader uses the precomputed time mappings to read the desired indices of events based on the start and end timestamps and loads them into memory. It also loads the corresponding grayscale images captured at the beginning and at the end of the event stream, along with the synchronized optical flow. The user can stride over the events using event indices, timestamps, or grayscale image indices. The data reader goes hand in hand with the included data iterators, which allow users to sequentially stride over the data. Such functionality allows the user to obtain event-based data sequentially, without the need for manual indexing.

        \emph{eWiz} also aims to support multiple \gls*{ebc} models to ensure compatibility with publicly available datasets. Several data converters are available for transforming these data types into the \emph{eWiz} format, including the DAVIS346 \citep{davis346} and Prophesee \citep{prophesee} cameras, along with popular datasets such as MVSEC \citep{mvsecZhu18} and DSEC \citep{dsecGehrig21}.
    
    \subsection{Data Encoding and Augmentation}
        Traditionally, \gls*{cnn}-based approaches \citep{evflownetZhu18, egoflownetZihao19, biflowWan22} require the \gls*{ebc} output to be converted from raw streams of event data to two-channel event images, where one channel represents positive polarities and the other represents negative polarities. To support this, \emph{eWiz} provides different encoders. Currently, two encoding schemes are implemented: Gaussian \citep{gaussianDing22} and event counting \citep{countNguyen19}. The former computes pixel values by summing each event’s polarity weighted by a normalized Gaussian kernel based on the event timestamp and a scaling factor $\lambda$ for each bin. The latter simply groups events into fixed time intervals, with pixel values computed by summing the polarities of all events within each bin.
    
        \emph{eWiz} also provides numerous event-based temporal and spatial data augmentations, including techniques such as time warping, noise injection, event flipping, and spatial transformations, all of which are fully compatible with PyTorch and Tonic. These augmentations enable enhanced model generalization by simulating diverse real-world conditions, making them ideal for training robust event-based neural networks.

    \subsection{Data Visualization}
        \emph{eWiz} includes several rendering modules for visualizing the acquired data. The primary renderer replays captured sequences, overlaying event images on top of grayscale images for an intuitive view of the data. Additionally, the library offers separate visualizers for event-based data, grayscale images, and optical flow, with the capability to save rendered sequences. Users can visualize the data as video sequences, individual (encoded) frames, or 3D volumes of event streams with interleaved frames.

    \subsection{Algorithms}
        \emph{eWiz} not only supports data processing for \gls*{nn}-based approaches but also supports algorithms, such as contrast maximization \citep{mcShiba2022}, that operate directly on raw event streams. Contrast maximization involves warping the event stream based on a proposed optical flow, iteratively optimizing the optical flow field to maximize the contrast computed between the warped and original event streams. Higher contrast, indicated by well-defined edges, signals improved alignment of events. As such, \emph{eWiz} also provides warping tools that enable the implementation of various motion models based on the camera movement type.

    \subsection{Loss Functions and Evaluation Metrics}
        \emph{eWiz} supports multiple loss functions for training \glspl*{cnn} and \glspl*{snn}. For instance, it supports the photometric loss \citep{evflownetZhu18} as a self-supervised approach to minimize pixel-wise intensity differences between two grayscale images captured immediately before and after the input events stream. The predicted flow field is used to warp the grayscale images and optimize alignment. This loss can be combined with Charbonier loss \citep{charbonnierSun14}, a smooth approximation of the L1 loss that is less sensitive to outliers, as well as a smoothness loss serving as a regularizer to address the aperture problem and ensure continuous flow across sparse event-based data. In addition, the library includes losses specifically designed for the contrast maximization algorithm \citep{mcGallego2018, mcGallego2019} to evaluate warped event streams. These include image variance and gradient magnitude loss functions, along with the improved variants presented by \citet{mcShiba2022}, where the losses are normalized and applied in a multi-focal manner. To further enhance performance, \emph{eWiz} also provides a regularizer that smooths the predicted flow across the image, helping address the sparsity of event-based data and ensuring continuous flow, ultimately improving convergence during training.

        Moreover, \emph{eWiz} can be used not only to evaluate the performance of trained models but also to track their performance during training. Currently, the library supports performance metrics for optical flow prediction tasks, such as the \gls*{aee} \citep{evflownetZhu18}. AEE is defined as the distance between the endpoints of the predicted and ground-truth flow vectors, while the AAE measures the angular difference between the predicted and ground-truth flows. Both metrics are averaged over the number of pixels. Due to the sparsity of event-based data, these metrics are only computed for pixels where events have occurred. Another implemented metric is the outlier percentage, which refers to the percentage of pixels in the image with an endpoint error exceeding a specified threshold.


\section{Dataset Structure}
\label{sec:dataset_structure}
    The \emph{eStonefish-Scenes} dataset follows the \emph{eWiz} format, which stores data in HDF5 files \citep{hdf5Format}. Specifically, \texttt{events.hdf5} holds the event streams, \texttt{gray.hdf5} contains grayscale image sequences associated with each event stream, and \texttt{flow.hdf5} stores the optical flow information. Metadata and configuration details, including sensor specifications and recording parameters, are saved in props.json. This structure facilitates straightforward access to each data type, improving data handling and compatibility with various processing pipelines. The directory structure is summarized below:
    \begin{figure}[!h]
        \centering
        \framebox[0.3\textwidth]{%
            \begin{minipage}{0.25\textwidth}
                \dirtree{%
                    .1 dataset\_root.
                    .2 events.hdf5.
                    .2 gray.hdf5.
                    .2 flow.hdf5.
                    .2 props.json.
                }
            \end{minipage}
        }
        \caption[Data Structure]{\centering Directory structure of the \emph{eStonefish-Scenes} data repository.}
        \label{fig:repo_structure}
    \end{figure}

\section{Ground Truth Acquisition}
\label{sec:ground_truth_acquisition}
    A crucial aspect of validating a synthetic dataset is evaluating the sim-to-real transferability \citep{surveyPitkevich24}. This is done in \secref{sec:sim-to-real_evaluation}, using a real-world dataset collected under controlled conditions in an indoor testing pool. The dataset is described in this section.

    \subsection{Experimental Setup}
        Data was collected using a BlueROV2 equipped with a downward-facing DAVIS346 event camera. The \gls*{rov} was deployed in the indoor testing pool at the Center for Research in Underwater Robotics and Vision (CIRS) in Girona. The ROV was manually controlled to perform six-degree-of-freedom motions, including translations along the forward–backward, lateral, and vertical axes, as well as yaw rotations.
        
        The DAVIS346 sensor recorded both asynchronous event streams and grayscale images at a resolution of 346×260 pixels with a \SI{70}{\degree} field of view. Eight sequences were recorded, each lasting 20 seconds. The pool was illuminated with natural ambient light. A high-resolution reef-like poster covering the entire pool floor was used to provide underwater textures, with the camera positioned at an average distance of 2.5 meters from the poster surface.

        \begin{figure}[!h]
            \centering
                \includegraphics[width=0.45\textwidth]{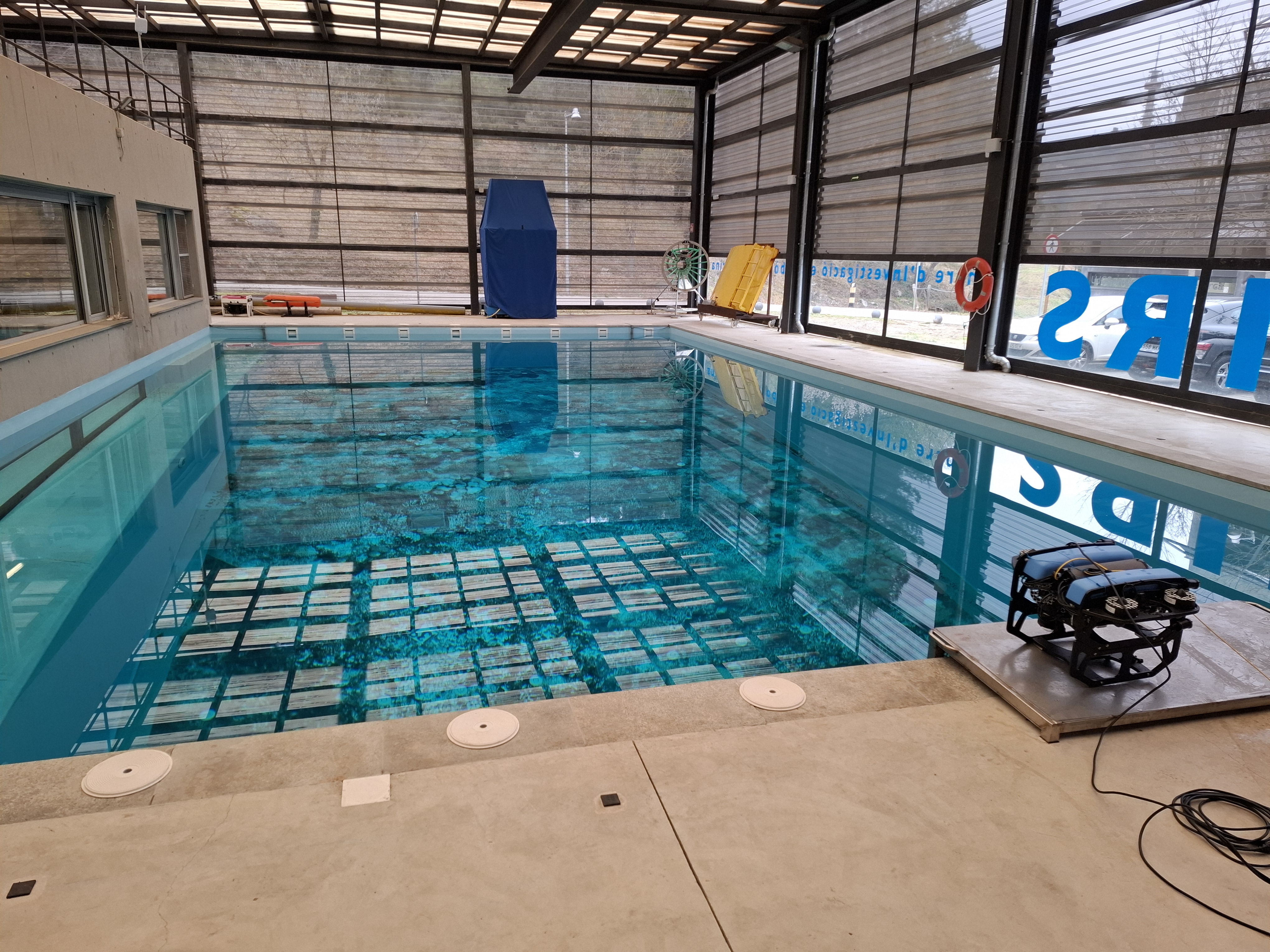}
                \caption{CIRS testing pool with floor poster used for ground-truth acquisition.}
            \label{fig:pool}
        \end{figure}

    \subsection{Ground Truth Flow Computation}
        Following data acquisition, we estimated ground-truth optical flow using homography-based registration of the grayscale frames to the poster in the pool.

        \subsubsection{Camera Calibration}
            The DAVIS346 calibration was performed using a checkerboard pattern, in front of which the camera was moved at different orientations. The grayscale images were then undistorted using the resulting intrinsics. This correction mitigated distortion-induced geometric errors during image-to-poster registration. The resulting flow fields were then reprojected into the undistorted domain to maintain consistency with the recorded event-based data.

        \subsubsection{Flow Computation}
            For each undistorted grayscale frame, SIFT features were extracted and matched to a fixed reference image of the poster. Using matched keypoints $\{(p_i^{\text{frame}}, p_i^{\text{poster}})\}$, we estimated a homography $H_{\text{image} \rightarrow \text{poster}}$ via RANSAC:
            \begin{equation}
                p_i^{\text{poster}} \sim {}^P H_i \cdot p_i^{\text{image}}
            \end{equation}

            For two consecutive frames $A$ and $B$, the relative homography was computed as:
            \begin{equation}
                {}^B H_A = {}^P H_B^{-1} \cdot {}^P H_A
            \end{equation}
    
            To generate dense optical flow, we defined a grid of pixel coordinates $\{x_i\}$ in frame $A$, and warped them using the relative homography:
            \begin{equation}
                \hat{x}_i = \frac{{}^B H_{A} \cdot [x_i, y_i, 1]^T}{z_i}, \quad f_i = \hat{x}_i - x_i
            \end{equation}
    
            This yields a dense flow field $f \in \mathbb{R}^{H \times W \times 2}$ between each pair of grayscale frames.

            \begin{figure}[!h]
                \centering
                    \includegraphics[width=0.55\textwidth, angle=270]{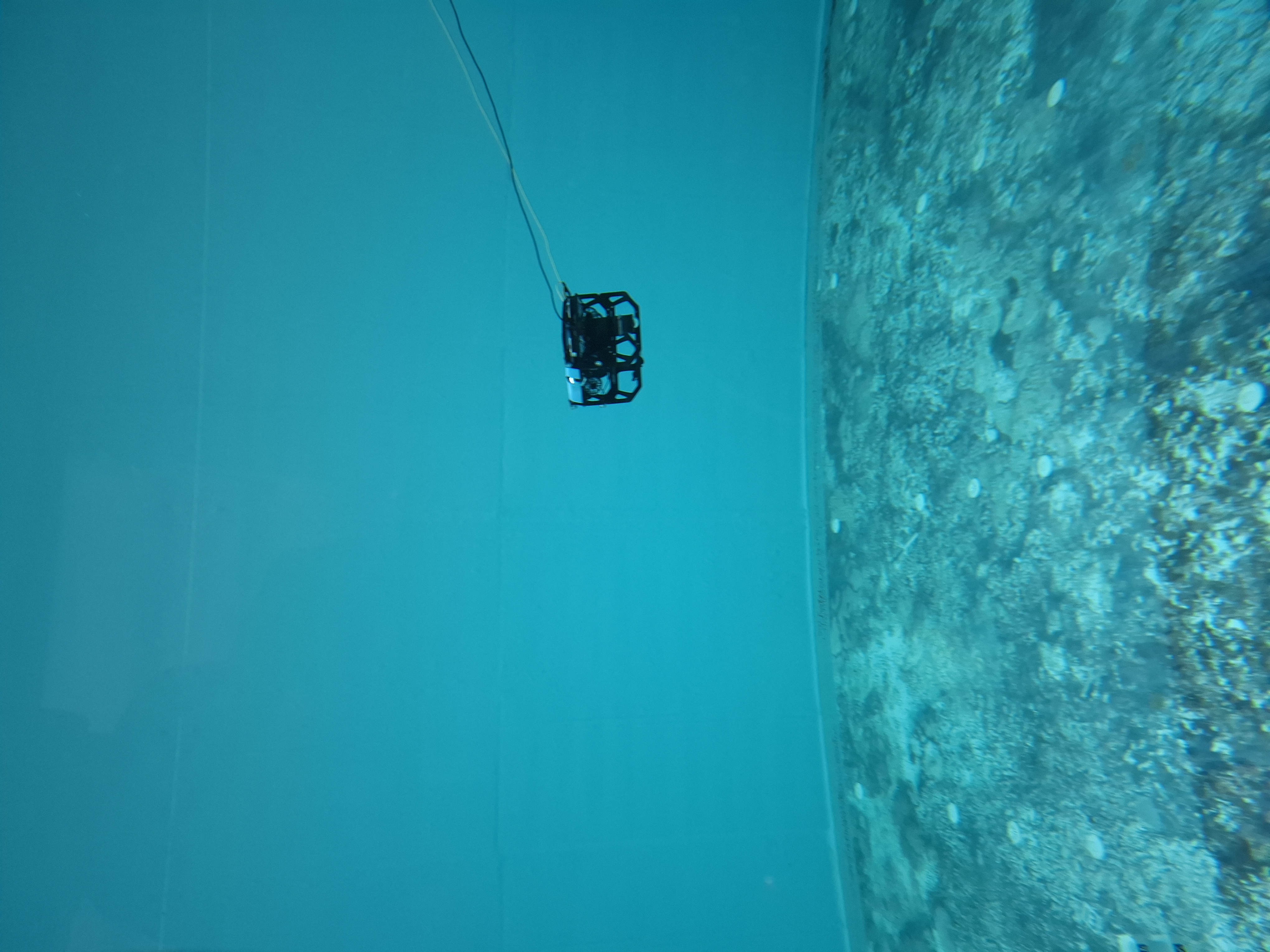}
                    \caption{ROV inside the testing pool during data acquisition.}
                \label{fig:pool_robot}
            \end{figure}

        \subsubsection{Outlier Rejection}
            During homography estimation, outlier rejection was performed using RANSAC. The inlier threshold was empirically set to 3 pixels based on the reprojection error between matched keypoints. For each frame, the proportion of inliers was computed as:
            \begin{equation}
                \eta = \frac{N_{\text{inliers}}}{N_{\text{total}}} \times 100
            \end{equation}
            
            where $N_{\text{inliers}}$ denotes the number of keypoints satisfying the reprojection constraint:
            \begin{equation}
            ||{}^P H_{i} p^{\text{image}}_i - p^{\text{poster}}_i|| < 3 \text{px}
            \end{equation}
            
            The distribution of $\eta$ was analysed across all frames in each sequence to assess registration robustness. Histograms of these values indicated that, for the majority of frames, the outlier percentage $(100 - \eta)$ remained below 30\%, confirming the stability of the chosen threshold across the dataset.

    \subsection{Uncertainty Estimation}
        To estimate the uncertainty in the homography-based optical flow, we used a Monte Carlo method based on keypoint reprojection errors. For each frame, SIFT keypoints from the reference poster image were matched to the current frame, and a homography $H_{\text{poster} \rightarrow \text{image}}$ was estimated using RANSAC. The reprojection error was computed by projecting each poster keypoint into the image frame and comparing it with its corresponding detected keypoint:
        \begin{equation}
            \epsilon_i = \left\| H_{\text{poster} \rightarrow \text{image}} \cdot p_i^{\text{poster}} - p_i^{\text{image}} \right\|
        \end{equation}

        The reprojection uncertainty for each frame was computed by aggregating all point-wise reprojection errors in both $x$ and $y$ directions into a single vector. Let the reprojection error for keypoint $i$ be $ \boldsymbol{\epsilon}_i = (\epsilon_{x, i}, \epsilon_{y,i})$. The overall noise level was estimated as the standard deviation of all concatenated components:
        \begin{equation}
            \sigma = \sqrt{\frac{1}{2N - 1} \sum_{i=1}^{N} \left[ (\epsilon_{x,i} - \bar{\epsilon}_x)^2 + (\epsilon_{y,i} - \bar{\epsilon}_y)^2 \right]}
        \end{equation}
        
        \noindent where $N$ is the number of matched keypoints. The resulting $\sigma$ values were then averaged across all frames to obtain a representative uncertainty estimate for the dataset. This value was used as the noise scale in a Monte Carlo simulation.
        
        For each of $N_s$ Monte Carlo iterations, we:
        \begin{enumerate}
            \item Added Gaussian noise to the poster keypoints:
                \begin{equation}
                    \tilde{p}_i = p_i^{\text{poster}} + \mathcal{N}(0, \sigma^2)
                \end{equation}
            \item Re-estimated the homography $\tilde{H}_{\text{poster} \rightarrow \text{image}}$ from noisy keypoints.
            \item Recomputed the dense optical flow $f_i^{(n)}$ from the resulting relative homography.
        \end{enumerate}

        Each pair of noisy homographies, $\tilde{H}_a$ and $\tilde{H}_b$, was used to warp a uniform grid of image coordinates and compute the corresponding optical flow sample $\mathbf{f}^{(n)} \in \mathbb{R}^{H \times W \times 2}$. Let $\mathbf{d}^{(n)}_{xy} = \mathbf{f}^{(n)}(x,y) - \bar{\mathbf{f}}(x,y)$ denote the deviation of the $n$-th Monte Carlo flow sample from the mean. The per-pixel covariance is then:
        \begin{equation}
            \mathbf{\Sigma}(x,y) = \frac{1}{N_s} \sum_{n=1}^{N_s} \mathbf{d}^{(n)}_{xy} \left(\mathbf{d}^{(n)}_{xy}\right)^{\!\top}
        \end{equation}
        

        A value of $N_s = 20$ Monte Carlo iterations was used, based on preliminary testing that showed no noticeable differences in the covariance matrices beyond this number of iterations. This yields a $2 \times 2$ covariance matrix per pixel, encoding both the variance and correlation of the horizontal and vertical flow components. To maintain consistency with the raw event-based data, both the resulting flow fields and their associated covariance maps were distorted using the camera intrinsics before being stored in the dataset.

\section{Sim-to-real Evaluation}
\label{sec:sim-to-real_evaluation}
    Building on the real data acquisition, we now evaluate the capacity of models trained exclusively on the synthetic \emph{eStonefish-Scenes} dataset to generalise to real underwater data. This evaluation aims to quantify the sim-to-real transfer performance and assess how well the proposed dataset supports learning event-based optical flow representations under real imaging conditions.

    \subsection{Network Architecture}
\begin{figure*}[!h]
        \centering
            \includegraphics[width=1\textwidth]{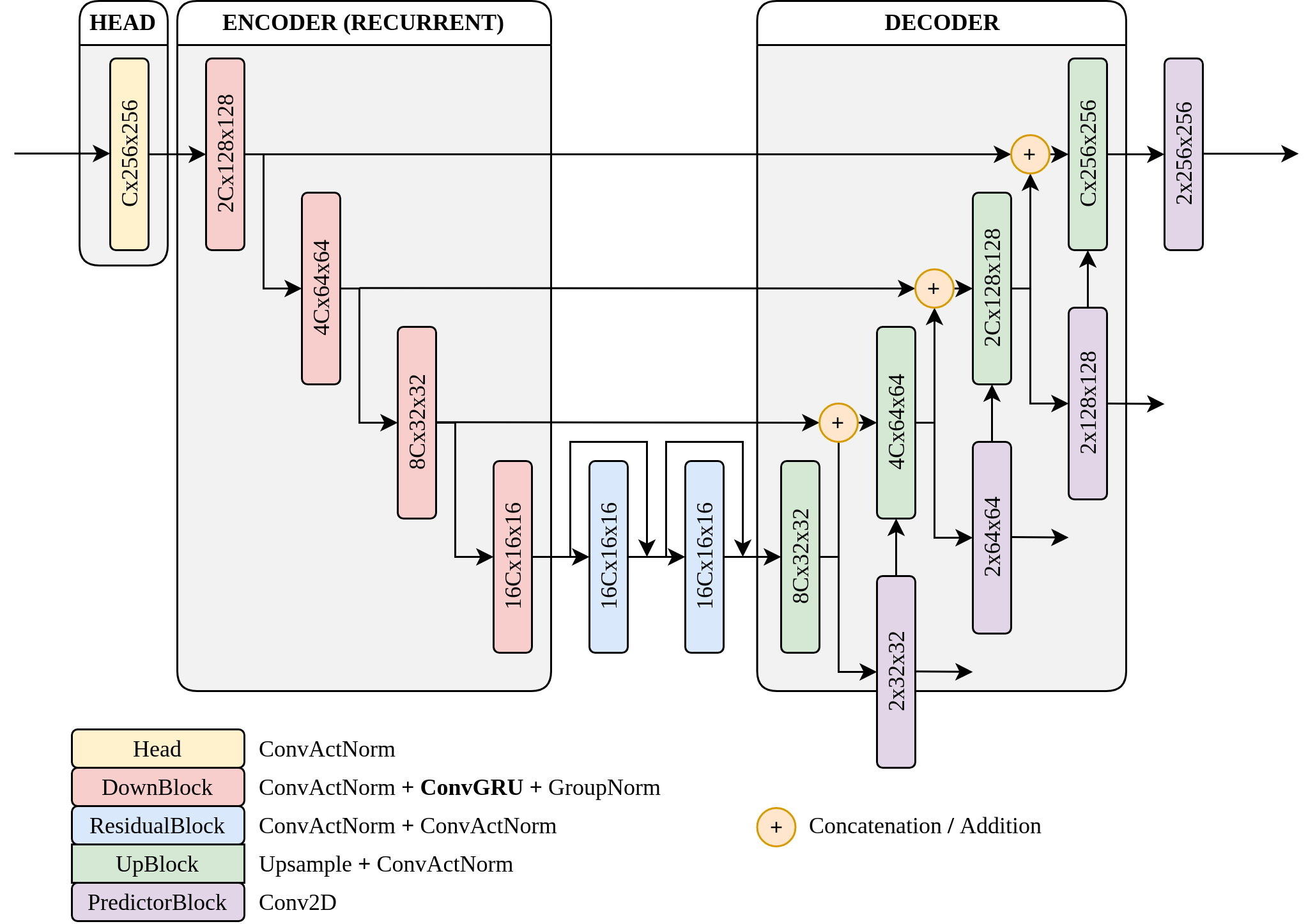}
            \caption{Architecture of the ConvGRU-based neural network used for optical flow prediction.}
        \label{fig:convgru_network}
        \end{figure*}

        The network architecture adopted in this work is inspired by the ConvGRU-based recurrent framework in \citep{firenetHagenaars21}. Similar to their design, our model is a time-driven convolutional recurrent network built around ConvGRU modules integrated within the encoder. A shallow convolutional head precedes the encoder, which consists of four downsampling blocks, each comprising a convolutional layer followed by group normalisation. The number of feature channels doubles after each down-sampling stage. The bottleneck section includes two residual blocks with skip connections. The decoder mirrors the encoder with four upsampling blocks that apply bilinear upsampling followed by a convolutional layer. Each upsampling stage includes a predictor block, composed of a single convolutional layer that outputs the optical flow at its respective resolution. A diagram of the network architecture is shown in \figref{fig:convgru_network}.

    \subsection{Dataset Preparation}
        For training, we employed the static sequences of the \emph{eStonefish-Scenes} dataset. Dynamic sequences were excluded to avoid inconsistencies in the smoothness loss caused by independent object motion. For evaluation, we used the real-world dataset, described in \secref{sec:ground_truth_acquisition}, to assess sim-to-real generalisation. The input event data were encoded using the Gaussian temporal encoding scheme proposed by \citep{encodingDing21}, which preserves both spatial and temporal event distributions.

        Input volumes were centre-cropped to $256 \times 256$ pixels. Spatial augmentations comprised random horizontal and vertical flips. Temporal robustness was encouraged by injecting synthetic event noise: for each network input, 200 additional events were added to the event stream. Training samples were formed by segmenting each sequence between every pair of consecutive grayscale frames. The network therefore received the events occurring strictly within $[t, t + \Delta t]$ for successive grayscale timestamps.

    \subsection{Multi-scale Loss \& Metrics}
    \label{sec:metrics}
        The network was trained using the self-supervised loss formulation introduced by \citet{evflownetZhu18}, combining a photometric reconstruction term with a spatial smoothness regulariser. The photometric component measures the intensity difference between the reference grayscale frame and the flow-warped target frame, while the smoothness term enforces continuity of the predicted flow field within local neighbourhoods. The latter assumption holds for static, rigid environments, which motivated the use of only static sequences for training. Training loss was computed using the Charbonnier penalty to mitigate the effect of outliers.

        Network performance was evaluated using the Average Endpoint Error (AEE) and the Average Angular Error (AAE), following standard definitions in the optical flow literature. In benchmarks such as KITTI, a pixel is typically considered an outlier when its endpoint error exceeds three pixels (or 5\% of the flow magnitude), and only the percentage of such outliers is reported. This fixed threshold assumes uniform reliability across all pixels, which is unsuitable for event-based data where uncertainty varies spatially. To account for this, each pixel in our evaluation was assigned a weight $w_{xy}$ derived from the inverse of the area of its uncertainty ellipse computed from the per-pixel covariance matrix. The weight was clipped such that $w_{xy} \le 1$. The weighted AEE was then defined as:
        \begin{equation}
            \mathrm{AEE}_{w} = \frac{\sum_{x, y} w_{xy}\sqrt{(u_{xy}-u^{gt}_{xy})^{2}+(v_{xy}-v^{gt}_{xy})^{2}}}{\sum_{x,y} w_{xy}}
        \end{equation}

        The weighted AAE was computed as:
        \begin{equation}
            \mathrm{AAE}_{w} = \frac{\sum_{x, y} w_{xy}\arccos\left(\frac{u_{xy}u^{gt}_{xy}+v_{xy}v^{gt}_{xy}+1}{\sqrt{(u_{xy}^{2}+v_{xy}^{2}+1)({u^{gt}_{xy}}^2+{v^{gt}_{xy}}^2+1)}}\right)}{\sum_{x,y} w_{xy}}
        \end{equation}
        
        This formulation integrates the spatially varying confidence of the optical flow, attenuating the contribution of uncertain regions while preserving global consistency. It therefore provides a more representative measure of performance in real-world, noise-prone event-based scenarios.

    \subsection{Training and Evaluation}
        Our model was trained on an NVIDIA Quadro RTX 6000 GPU with 24 GB VRAM for 200 epochs with a batch size of 16. We used the AdamW optimiser with a weight decay of $0.01$ and learning rate of $0.004$, decayed using a polynomial learning rate scheduler with a warm-up of $3$ epochs. The models were implemented in PyTorch $2.9.1$ and Python $3.8$. The source code with all hyperparameter configurations and pre-trained models will be made available at ~\url{https://github.com/CIRS-Girona/ebof}.

        The model was then evaluated on a standard laptop equipped with an NVIDIA RTX 3070 GPU running Ubuntu 20.04, Python $3.8$ and PyTorch $2.8.1$. For validation, the network was tested on a held-out subset corresponding to 10\% of the synthetic \emph{eStonefish-Scenes} dataset, comprising \textbf{607} input samples, as well as on the real-world dataset, comprising \textbf{5,984} input samples. No fine-tuning was performed on the real data. Performance was assessed using the uncertainty-aware Average Endpoint Error (AEE) and Average Angular Error (AAE) metrics described in \secref{sec:metrics}.

    \subsection{Results}
        \subsubsection{Synthetic Data Evaluation}
            We first evaluate the performance of the proposed ConvGRU-based network on a held-out subset of the synthetic \emph{eStonefish-Scenes} dataset. This subset corresponds to 10\% of the available synthetic data and comprises \textbf{607} input samples, each formed by encoding the events occurring between two consecutive grayscale frames. The evaluation set includes both down-looking and front-looking camera configurations and covers a range of motion patterns, including forward and backward translation, vertical ascent and descent, and combined translational and rotational motion.

\begin{figure*}[!h]
                \centering
                    \includegraphics[width=1\textwidth]{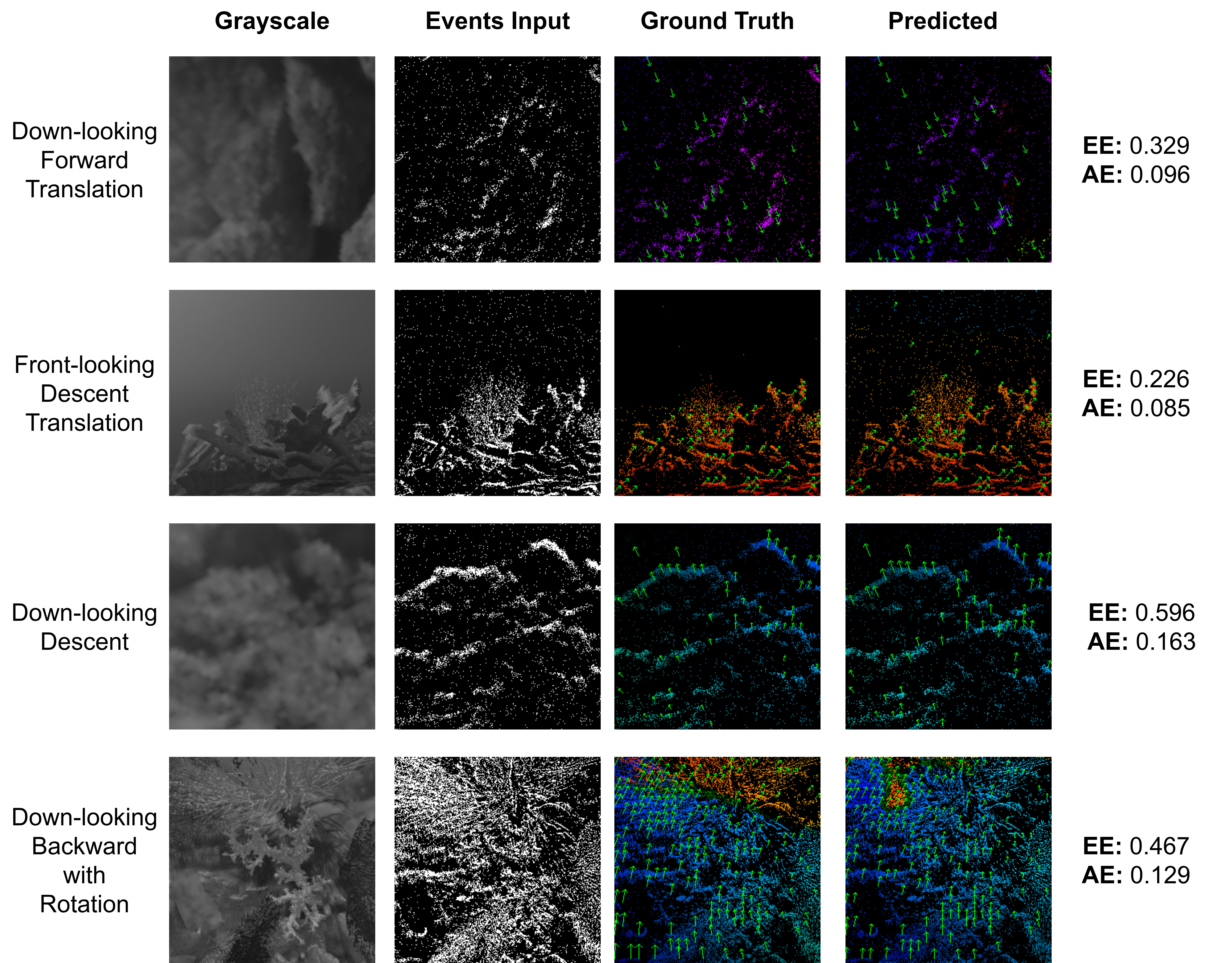}
                    \caption{Qualitative results on the synthetic eStonefish-Scenes test set. Each row corresponds to a different camera configuration and motion pattern. Columns show (from left to right) the reference grayscale image, the encoded event input accumulated between two consecutive grayscale frames, the ground-truth optical flow, and the predicted optical flow produced by the ConvGRU-based network. Optical flow is visualised using a colour-coded representation with arrows indicating flow direction and magnitude. The Endpoint Error (EE) and Angular Error (AE) for each example are reported on the right. The results demonstrate that the network accurately captures dominant translational motion and preserves coherent flow directions across diverse synthetic underwater scenarios.}
                \label{fig:synthetic_qualitative}
            \end{figure*}

            \figref{fig:synthetic_qualitative} presents representative qualitative results from the synthetic test set. For each example, we show the reference grayscale frame, the encoded event input, the ground-truth optical flow, and the predicted flow. The first row illustrates a down-looking forward translation, where the predicted flow closely matches the ground truth in both magnitude and direction, achieving an endpoint error of $EE = 0.329$ and an angular error of $AE = 0.096$. The second and third rows demonstrate front-looking and down-looking descent motions, respectively. In these cases, the network accurately recovers the dominant vertical flow components induced by heave, although small discrepancies appear in regions with sparse events or low texture. The final example combines backward translation with rotational motion, producing more complex flow patterns. While the predicted flow captures the overall structure of the motion, local deviations are more pronounced. Across all examples, the model preserves coherent flow directions in regions with sufficient event activity. These qualitative results corroborate the quantitative findings and highlight the network's ability to generalise across diverse synthetic underwater scenarios.

        \subsubsection{Real Data Evaluation}
\begin{figure*}[!h]
                \centering
                    \includegraphics[width=1\textwidth]{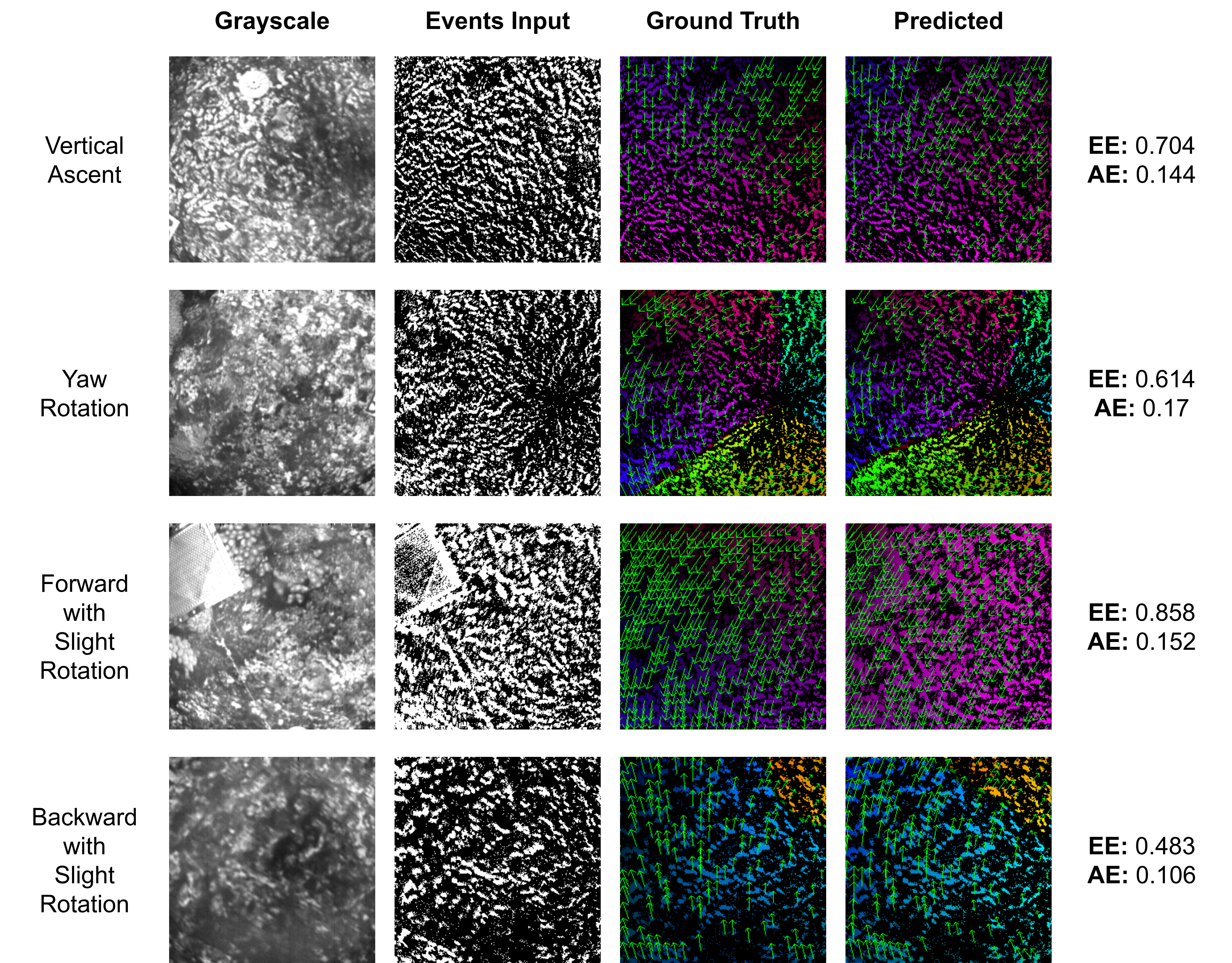}
                    \caption{Qualitative results on the \emph{eStonefish-Scenes} real-world validation dataset. Each row corresponds to a different ROV motion pattern acquired using a DAVIS346 event camera mounted on a BlueROV2 platform. Columns show (from left to right) the reference grayscale image, the encoded event input accumulated between consecutive grayscale frames, the ground-truth optical flow computed via frame-to-poster homography registration, and the predicted optical flow produced by the ConvGRU-based network. The results demonstrate strong qualitative agreement between prediction and ground truth across diverse real underwater scenarios, despite the network being trained exclusively on synthetic data.}
                \label{fig:real_qualitative}
            \end{figure*}

            \figref{fig:real_qualitative} presents qualitative results obtained on the real-world dataset. Each example shows the reference grayscale image, the encoded event input, the ground-truth optical flow computed via frame-to-poster registration, and the corresponding network prediction. Across all examples, the predicted flow fields exhibit strong qualitative agreement with the ground truth in regions with sufficient event activity, accurately capturing dominant motion directions and flow structure. Discrepancies are primarily observed near textureless regions, depth discontinuities, or areas affected by registration uncertainty, where the underlying ground truth is itself less reliable. Notably, the network preserves coherent flow patterns even under complex motion involving rotation, demonstrating robustness to real-world sensing artefacts despite being trained exclusively on synthetic data.

        \subsubsection{Quantitative Evaluation}
            \begin{table*}[!h]
                \centering
                \begin{tabular}{lcc}
                    \toprule
                    Dataset & AEE (Pixels) & AAE (Radians) \\
                    \midrule
                    Synthetic \emph{eStonefish-Scenes} (Test set, 10\%) & 0.636 & 0.173 \\
                    Real \emph{eStonefish-Scenes} & 0.79 & 0.2 \\
                    \bottomrule
                \end{tabular}
                \caption{Quantitative evaluation results on the synthetic and real-world \emph{eStonefish-Scenes} datasets. AEE is measured in pixels and AAE in radians.}
                \label{table:mixed_quantitative}
            \end{table*}
    
            \tabref{table:mixed_quantitative} reports quantitative performance on both the synthetic \emph{eStonefish-Scenes} test set and the real-world dataset. For the synthetic data, where dense ground-truth optical flow is available from simulation, performance is evaluated using the unweighted AEE and AAE. Under these conditions, the proposed ConvGRU-based model achieves an average endpoint error of 0.636 pixels and an average angular error of 0.173 radians.

            For the real-world dataset, where ground truth is obtained via homography-based registration and is inherently affected by calibration and feature-matching uncertainty, performance is evaluated using uncertainty-aware metrics. In this setting, the model attains an uncertainty-weighted endpoint error of 0.79 pixels and an uncertainty-weighted angular error of 0.2 radians without any fine-tuning on real data. The observed performance gap between synthetic and real data reflects the expected sim-to-real discrepancy induced by sensor noise, optical distortions, and imperfect ground-truth estimation. Importantly, incorporating uncertainty into the evaluation attenuates the influence of ambiguous regions and yields more reliable performance estimates on real underwater data.

\section{Conclusion}
\label{sec:conclusion}
    This work addresses the challenge of developing reliable optical flow prediction networks for underwater environments, where collecting large-scale, annotated, event-based optical flow datasets is exceptionally difficult due to the complexities of underwater imaging. To overcome this, we introduce the \emph{eStonefish-Scenes} dataset, along with a user-friendly pipeline to extend and adapt it. To the best of our knowledge, no prior event-based datasets exist for underwater applications. The dataset incorporates scenes capturing event-based data from an \gls*{rov} navigating a dynamic, coral-rich environment filled with diverse marine flora and fauna. We also introduce \emph{Stonefish-SceneGen}, an automated scene generator that eliminates the tedious and time-consuming process of manually creating reef environments. Additionally, we present \emph{Stonefish-Boids}, a package that simulates realistic schooling behaviour of fish. The dataset and supporting packages were created with the goal of accelerating research on \glspl*{ebc} for underwater applications, especially for training \glspl*{cnn} and \glspl*{snn}. We also present \emph{eWiz}, a comprehensive event-based data manipulation library that includes an optimised event data format. The library features data converters, writers, readers, and processing modules focused on data management, compression, and optimisation. Additionally, it offers various loss functions, performance metrics, and visualisers to facilitate work with these datasets, along with an implementation of the contrast maximisation algorithm. Both \emph{eWiz} and \emph{eStonefish-Scenes} are designed to streamline research on \glspl*{ebc} and motion estimation for autonomous underwater vehicles by providing easy-to-use libraries for generating, processing, and evaluating event-based datasets. In the long run, \emph{eWiz} will support not only optimisation-based algorithms but also deep learning-based and \gls*{snn}-based alternatives. We plan to add more functionalities, such as ground truth generation via optimisation and \gls*{snn}-based training algorithms.

    Beyond dataset generation, we validated the practical utility of \emph{eStonefish-Scenes} through a real-world experimental study using an event-based camera mounted on a BlueROV2 platform. Ground-truth optical flow was computed via homography-based frame-to-poster registration, and per-pixel uncertainty was explicitly estimated and incorporated into the evaluation metrics. A ConvGRU-based optical flow network trained exclusively on synthetic data demonstrated effective sim-to-real transfer when evaluated on real underwater sequences, with qualitative and quantitative results showing strong agreement in dominant motion patterns despite the absence of real-data fine-tuning. The combined quantitative and qualitative results demonstrate that the proposed synthetic dataset supports effective learning of transferable event-based motion representations for underwater environments. The use of uncertainty-aware evaluation metrics provides a more reliable assessment of performance in real-world settings, where ground truth is inherently imperfect. By downweighting regions with high uncertainty, the proposed evaluation framework avoids over-penalising predictions in ambiguous areas and offers a principled alternative to fixed outlier thresholds commonly used in optical flow benchmarks. These results suggest that synthetic event-based datasets, when combined with realistic simulation, can substantially reduce the reliance on large-scale real-world data collection for underwater robotics applications.

    Future work will focus on extending the dataset to include dynamic scenes and independently moving objects, as well as exploring spiking neural network architectures to more fully exploit the temporal nature of event-based data.

\begin{funding}
    Jad Mansour was supported by the Joan Oró Grant no. 2024 FI-2 00762. This study was also supported in part by the Spanish government through the ASSIST project PID2023-149413OB-I00 and the IURBI project CNS2023-144688.
\end{funding}

\begin{dci}
    The authors declared no potential conflicts of interest with respect to the research, authorship, and/or publication of this article.
\end{dci}

\begin{credit}
    \textbf{Jad Mansour:} Conceptualization, Methodology, Software, Data Curation, Writing - original draft; \textbf{Sebastian Realpe:} Software, Writing - original draft; \textbf{Hayat Rajani:} Conceptualization, Methodology, Supervision, Validation, Writing - original draft; \textbf{Michele Grimaldi:} Software, Writing - review \& editing; \textbf{Rafael Garcia:} Methodology, Project Administration, Supervision, Writing - review \& editing; \textbf{Nuno Gracias:} Project Administration, Supervision, Writing - review \& editing.
\end{credit}

\bibliographystyle{SageH}
\bibliography{references.bib}

\end{document}